\documentclass[runningheads]{llncs}

\usepackage{accv}

\usepackage{accvabbrv}

\usepackage{graphicx}
\usepackage{booktabs}

\usepackage[accsupp]{axessibility}  % Improves PDF readability for those with disabilities.

\usepackage[pagebackref,breaklinks,colorlinks,citecolor=accvblue]{hyperref}
\usepackage{orcidlink}

\begin{document}

% ---------------------------------------------------------------
% TODO REVIEW: Replace with your title
\title{Continual Learning Improves Zero-Shot \\ Action Recognition}

% TODO REVIEW: If the paper title is too long for the running head, you can set
% an abbreviated paper title here. If not, comment out.
%\titlerunning{Abbreviated paper title}

% TODO FINAL: Replace with your author list. 
% Include the authors' OCRID for the camera-ready version, if at all possible.
\author{Shreyank N Gowda\inst{1} \and
Davide Moltisanti\inst{2} \and
Laura Sevilla-Lara\inst{3}}

% TODO FINAL: Replace with an abbreviated list of authors.
\authorrunning{S.N. Gowda et al.}
% First names are abbreviated in the running head.
% If there are more than two authors, 'et al.' is used.

% TODO FINAL: Replace with your institution list.
\institute{University of Nottingham, UK \and
University of Bath, UK \and
University of Edinburgh, UK\\
\email{Shreyank.Narayanagowda@nottingham.ac.uk, dm2460@bath.ac.uk, l.sevilla@ed.ac.uk}}

\maketitle

\setcounter{footnote}{0}  % necessary to fix a bug in the .tex class (footnotes don't start from 1)

\begin{abstract}
Zero-shot action recognition requires a strong ability to generalize from pre-training and seen classes to novel unseen classes. Similarly, continual learning aims to develop models that can generalize effectively and learn new tasks without forgetting the ones previously learned. The generalization goals of zero-shot and continual learning are closely aligned, however techniques from continual learning have not been applied to zero-shot action recognition. In this paper, we propose a novel method based on continual learning to address zero-shot action recognition. This model, which we call {\em Generative Iterative Learning} (GIL) uses a memory of synthesized features of past classes, and combines these synthetic features with real ones from novel classes. The memory is used to train a classification model, ensuring a balanced exposure to both old and new classes. Experiments demonstrate that {\em GIL} improves generalization in unseen classes, achieving a new state-of-the-art in zero-shot recognition across multiple benchmarks. Importantly, {\em GIL} also boosts performance in the more challenging generalized zero-shot setting, where models need to retain knowledge about classes seen before fine-tuning.
\end{abstract}

% \begin{center}
%     \centering
%     \captionsetup{type=figure}
% \includegraphics[width=\textwidth]{teaser_gil_final_dm.pdf}
%     \caption{In standard zero-shot action recognition (left), a classifier is pre-trained on a large dataset, then fine-tuned and tested on new classes. The continual learning approach (middle-left) adds a memory replay to train on both old and new classes, improving the results in the zero-shot setting (middle-right) as well as on the pre-trained dataset (right) performance.%, thus improving on both new (top-right) and pre-trained dataset results (bottom right).
%     }
%     \label{fig:teaser}    
% \end{center}

\section{Introduction}
\label{sec:intro}

Traditional supervised deep learning has shown prowess in diverse applications~\cite{resnet,detr,smart,li2023uniformer,colornet}, but its strong dependence on large amounts of labeled data~\cite{imagenet,coco,i3d} 
has become its Achilles' heel, 
especially when encountering novel data categories. 
Zero-shot learning (ZSL) opens up new possibilities, focusing on classifying data from novel categories that are unseen during training. 
Continual learning (CL) is another area of research that addresses the same fundamental challenges involved in novel data, though in a distinct way: learning new tasks (or data) without forgetting previous knowledge. Indeed, ZSL and CL share the objective of building representations with high generalization abilities. In ZSL, the goal is to generalize from pre-train and seen classes to unseen classes, while in CL the goal is to generalize from a task to the next one without losing performance in the first one. Previous work has applied CL to improve ZSL in the image domain~\cite{cl1,cl2,cl3,cl4,cl5}, however, to the best of our knowledge, no previous work has leveraged CL for zero-shot action recognition in video.

In this work, we thus propose the first method for zero-shot action recognition based on a continual learning paradigm. We name our method \textbf{GIL}: \textbf{G}enerative \textbf{I}terative \textbf{L}earning.
GIL consists of three stages: during initialization we build a Replay Memory storing prototypical video representations for all classes in a pre-training dataset. A Generative Adversarial Network (GAN) is also trained to produce features resembling those stored in the memory. In the incremental learning stage we then gradually fine-tune the video model on a fine-tuning dataset, mixing features generated with the GAN from the memory (old knowledge) with real features obtained for the new classes (new knowledge). The incremental learning stage is alternated with the update stage, where we refresh the memory adding prototypical representations from the new classes. This cycle ensures that the model learns new classes while not forgetting the old ones, which is the basis of continual learning. GIL has a strong generalization ability and improves recognition both in the standard zero-shot setting and the more challenging generalized zero-shot setting (GZSL), where the model is tested also on classes seen during pre-training. This is demonstrated on three standard ZSL/GZSL action recognition benchmarks (UCF-101~\cite{ucf101}, HMDB-51~\cite{hmdb}, Kinetics-600~\cite{i3d,er}), where we improve by a wide margin of up to 20\%. Project page: \url{https://sites.google.com/view/gil-accv}.

% Our work is summarized in Figure~\ref{fig:teaser}. We deviate from standard zero-shot and incorporate a continual learning paradigm to learn stronger and more generalizable representations, which is beneficial in both ZSL and GZSL settings.

\section{Related Work}
\label{sec:related}

\noindent{\bf Zero-Shot Learning in Action Recognition.} 
Previous research focuses on establishing a shared embedding space between video features and semantic labels~\cite{xu2016multi,xu2017transductive}. Others explore error-correcting codes~\cite{qin2017zero}, pairwise class relationships~\cite{gan2016concepts}, and interclass dynamics~\cite{gan2015exploring}. Other approaches include leveraging out-of-distribution detectors~\cite{od} and graph neural networks~\cite{gao2019know}. Recently, clustering of joint visual-semantic features was proposed~\cite{claster}. This cluster-based technique and others, like ReST~\cite{rest} and JigSawNet~\cite{jigsaw}, emphasize the joint modeling of visual and textual features, albeit with varying approaches to bridge the visual and text spaces. Image-based foundational models have also shown to be excellent zero-shot learners in video. These models have been ported to videos with multi-modal prompting~\cite{vita-clip}, self-regulating prompts~\cite{srp} or by using cross-frame attention mechanism that explicitly exchanges information across frames~\cite{florence}. Enhancing semantic embeddings from simple word2vec~\cite{w2v} to more elaborate definitions~\cite{er} has shown great promise for ZSL. For instance, action ``Stories''~\cite{stories} enhance the semantic space by breaking down actions into descriptive steps. 
We use semantic embeddings to build semantic class representations. We employ Stories~\cite{stories} due to its good performance, 
and show that our method improves over the baseline model regardless of the semantic embedding we choose.
We also evaluate the more challenging generalized zero-shot learning (GZSL) setting. Recent work~\cite{chen2021semantics} uses a CVAE to disentangle visual features into semantic-consistent and unrelated parts, enhancing generalization.~\cite{han2021contrastive} combines feature generation with contrastive embeddings to create a more discriminative space for both real and synthetic sample.~\cite{chen2023zero} employs adversarial training with augmented samples to maintain semantic consistency and robustness. Our method differs from these by incorporating continual learning with a replay memory to combine synthesized features of past classes with real features of new ones. This updates knowledge incrementally, 
unlike the other methods that focus on static feature manipulation~\cite{chen2021semantics,han2021contrastive} or adversarial training~\cite{chen2023zero}.

\noindent{\bf Continual Learning.} Traditional deep learning models suffer from catastrophic forgetting~\cite{mccloskey1989catastrophic}. 
Continual learning aims at training models that can learn from a continuous stream of data without forgetting previously acquired knowledge~\cite{parisi2019continual}. 
Various methods have been proposed to tackle this. Some have explored the concept of elastic weight consolidation~\cite{kirkpatrick2017overcoming} to regularize the changes in network weights. Others have utilized memory augmentation strategies, like the use of external memory modules~\cite{graves2016hybrid}. Progressive neural networks~\cite{rusu2016progressive} expand the network architecture to leverage prior knowledge via
lateral connections to previously learned features to alleviate catastrophic forgetting. There are also methods that leverage meta-learning principles, allowing the model to learn how to learn across tasks~\cite{finn2017model}. Recent work has focused on distilling knowledge from one part of the model to another~\cite{hung2019compacting}, and generating instance level prompts~\cite{jung2023generating}. 
Another promising avenue is the use of generative replay~\cite{replay}, where generated samples from previous tasks are mixed with new data to reduce forgetting. In this work we use a feature generator network to build a replay memory~\cite{replay}. Instead of generating instance-like features, we generate class-like prototypical representations, which enhances efficiency. 
In class-incremental learning efficiency has also been investigated managing memory.~\cite{zhou2023model,zhou2024class} suggest  that focusing on different network layers, particularly the deeper ones, can improve memory efficiency and performance. This aligns with our approach, where we fine-tune the last two layers to leverage adaptability while maintaining core model stability.

\noindent{\bf Continual Learning and Zero-Shot Learning.} As discussed before, CL and ZSL share the goal of learning generalizable representations.
CL has been explored for ZSL in image tasks~\cite{skorokhodov2020class,cl4,cl5}. Most of these works convert the problem of standard zero-shot learning into that of a continual learning problem dubbed continual zero-shot learning (CZSL). CZSL mimics human lifelong learning by continuously incorporating new classes from the unseen world, evaluating the model on both seen and unseen categories. To tackle this, prior work has proposed class normalization~\cite{skorokhodov2020class}, semantic guided random walks~\cite{cl5}, experience replay with dark knowledge distillation~\cite{gautam2022tf} and meta-learned attributes~\cite{verma2024meta}. Most similar to our work is the use of generative replay for CZSL~\cite{cl4} that uses attribute information to generate synthetic features. In contrast, we store class prototype and associated distribution as noise and generate synthetic data using this. Further, CL has not been used for zero-shot action recognition in videos, and catastrophic forgetting has not been evaluated in this context. We propose the first study of this problem in zero-shot action recognition and propose the first method for zero-shot action recognition based on continual learning.

\section{Methodology}

\subsection{Zero-Shot Problem Formulation}

Let $P$ and $S$ be respectively a pre-training and a fine-tuning dataset for seen classes. 
Each element in $P$ and $S$ is a tuple $\left( x, y, a \right)$. Here, $x$ denotes the spatio-temporal features of a video; $y$ is the class label from the set of seen classes $Y_{S}$; and $a$ is the semantic representation of class $y$. This representation can be manually annotated or derived automatically, e.g., with Stories~\cite{stories}.
Let $U$ be the collection of pairs $\left (u, a\right)$, where $u$ is a class from the set of unseen classes $Y_{U}$, and $a$ indicates its corresponding semantic representation. The seen $Y_S$ and unseen classes $Y_U$ have no common elements.
In the ZSL setting, given a video input the objective is to determine a class label from the unseen class set, represented as $f_{ZSL}:X\rightarrow Y_{U}$. In the Generalized Zero-Shot (GZSL) setting, given a video input, the goal is to identify a class label from the combined set of seen and unseen classes, expressed as $f_{GZSL}:X\rightarrow Y_{S}\cup Y_{U}$.

\subsection{Preliminaries}

We now describe the two main building blocks that our model is based on: feature generation from ZSL and replay memory from CL. \\

\noindent{\bf Feature Generation.} The standard feature generation approach~\cite{clswgan,od,ggm,cvae} learns to generate visual features for unseen classes using a GAN~\cite{gans}. Given these synthetic features, a classifier that takes in visual features is trained to predict unseen classes. 
More concretely, the GAN consists of a feature generator network ($\mathcal{F}$), a discriminator ($G$) and a projection network ($H$). Given the visual input $x_i$ and its class label $y_i$, the generator takes in the semantic embedding $a_i$ 
for the class label $y_i$ and some noise $z$, and produces a synthetic visual feature vector $\hat{x}_i$. The projection network maps the visual feature $x_i$ to an approximation of the semantic embedding $\hat{a}_i$. The discriminator's job is to distinguish the synthesized features from the real features. 
All these networks are trained jointly, where the generator and the discriminator compete in a mini-max game. The optimization function to train these networks is as follows:

\begin{equation}
\label{eq:joint}
\begin{gathered}
\mathcal{L}_D= \mathbb{E}_{(x,a)\sim p_{(x_{S}\times a_{S})}}[G(x,H(x))] \\ 
 - \mathbb{E}_{z\sim p_z}\mathbb{E}_{a\sim p_a}[(G(\mathcal{F}(a,z),a))] \\
 - \alpha \mathbb{E}_{z\sim p_z}\mathbb{E}_{a\sim p_a}\left[\left(\left\|\nabla_{\hat{x}} G(\mathcal{F}(a,z) )\right\|_2-1\right)^2\right],
\end{gathered}
\end{equation}

where $p_{(x_{S}\times a_{S})}$ is the joint empirical distribution of visual and semantic descriptors of the seen classes, $p_a$ is the empirical distribution of seen classes semantic embeddings, $p_z$ is the noise distribution and $\alpha$ is a penalty coefficient.

There are two additional losses, used to improve the quality of the generated features. These are the classification regularized loss  $\mathcal{L}_{CLS}$~\cite{clswgan} and the mutual information loss $\mathcal{L}_{MI}$ ~\cite{belghazi2018mutual}. Putting all losses together, the objective function that we minimize to train the vanilla pipeline is: 

\begin{equation}
\label{eq:overall}
\begin{gathered}
\min_\mathcal{F} \min_H \max_G \mathcal{L_D} + \lambda_{1}\mathcal{L}_{CLS}(\mathcal{F}) + \lambda_{2}\mathcal{L}_{MI}(\mathcal{F}).
\end{gathered}
\end{equation}

Once these networks are trained on the seen classes, the generator is used to synthesize visual features for the unseen classes. \\

\noindent{\bf Continual Learning.}
Formally, given a sequence of tasks ${T_1, T_2, \ldots, T_n}$, where each task $T_i$ consists of a dataset $D{i} = {(x_{i1}, y_{i1}), (x_{i2}, y_{i2}), \ldots}$, the objective of CL is to perform well at $T_i$ without losing accuracy on any of the tasks $T_{j}$, where $1<j<i$. For ZSL and GZSL, we can consider the pre-training set $P$ to be $T_1$, each set of seen classes that are gradually added to the training set to be $T_{2,..,N-1}$ and the unseen classes to be $T_N$. 
Replay Memory~\cite{replay} is a powerful technique in CL to mitigate catastrophic forgetting. The idea is to leverage generative models to generate or ``replay'' samples from past tasks while learning new ones, preserving the knowledge from earlier tasks.  
This is achieved by incorporating a replay buffer $B$, which stores a subset of samples from previous tasks. A generator $\mathcal{F}$ is trained to generate samples that are representative of those in $B$. 
For a new task $T_i$, the model is trained on the current task data $D_{i}$ as well as on the replayed samples generated by $\mathcal{F}$, which helps generalization ability and alleviates catastrophic forgetting. 

\subsection{Our Approach: Generative Iterative Learning (GIL)}
\label{subsec:GIL}

\begin{figure}[t]
    \centering \includegraphics[width=\textwidth]{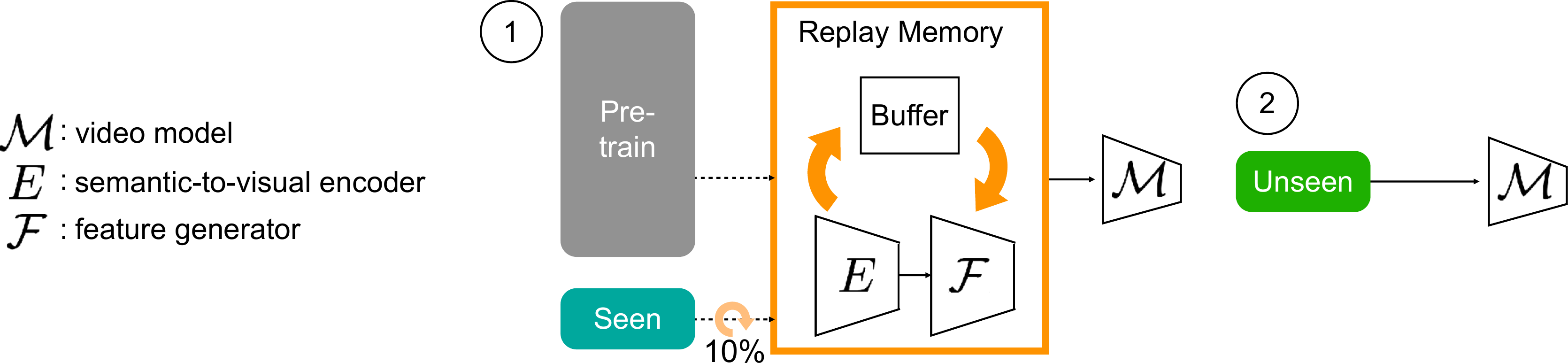}
\caption{GIL at a glance. The main idea is to use a Replay Memory~\cite{replay} from Continual Learning to fine-tune a video model $\mathcal{M}$. The Replay Memory consists of a buffer which contains classes from the pre-training and fine-tuning datasets. Samples in the buffer are generated with a semantic-to-visual encoder $E$ and a feature generator $\mathcal{F}$.}
\label{fig:coarse_GIL}
\end{figure}

A broad overview of our architecture can be seen in Fig.~\ref{fig:coarse_GIL}. 
The main idea is to use a Replay Memory to fine-tune a video model $\mathcal{M}$ which we will use for zero-shot prediction. The Replay Memory ensures that $\mathcal{M}$ does not lose its generalization abilities. The memory consists of a buffer, which contains data from the pre-training dataset as well as the seen classes from the fine-tuning dataset, which is incorporated progressively in small batches. The buffer starts off as real data and progressively includes synthetic data, generated by a conditional variational auto-encoder $E$ and a feature generation network $\mathcal{F}$. The model is trained iteratively, alternating the fine-tuning stage with a memory update stage until all classes in the fine-tuning dataset are seen.
We next describe each component in detail, and later describe how we train and test the framework. \\

\noindent{\bf Foundation Model.} 
Our approach is model-agnostic. We choose a foundation model $\mathcal{M}$ to encode the videos, so that they will live in a meaningful semantic space, which is essential in ZSL. 
We use X-Florence~\cite{florence} as our initial pre-trained model due to its good performance 
and will show results obtained with other backbones as well. X-Florence is a video model built on top of the image foundational model Florence~\cite{imgflorence}. 
X-Florence consists of three main components: a video encoder, a text encoder and a video-specific prompt generator that are pre-trained together (more details in~\cite{florence}). We only use the video encoder.
We adapt $\mathcal{M}$ by adding two fully-connected layers at the end that generate a video representation in the same space as the textual representation, which enables testing with a nearest neighbor search. We keep the backbone of $\mathcal{M}$ fixed at all times and only train these additional layers when fine-tuning $\mathcal{M}$.
\\

\noindent{\bf Class Prototype and Noise.} Models based on Replay Memory use a buffer that stores a subset of samples from each task. However, a random subset of samples does not necessarily reflect the true distribution of a class. Instead, we use class prototypes, which are the average feature representation ($\mu$) of each sample ($x_{i}$) in a class ($C$). 
We also use the distance between a class prototype and each sample (the standard deviation $\sigma(C)$) to better approximate the class distribution.  
\\

\noindent{\bf Replay Memory.} The Replay Memory~\cite{replay} consists of a buffer and a Conditional Variational Autoencoder (CVAE)~\cite{cvae}, which acts as a semantic-to-visual encoder. For each class seen in training we archive its prototype and noise in the buffer.
The CVAE receives a semantic embedding 
and is trained to output a latent variable that is equivalent to the concatenation of the prototype $\mu(C)$ and noise $\sigma(C)$, for all classes in the buffer. We denote the CVAE as the model $E$. Finally, we use an off-the-shelf feature generator model~\cite{clswgan} $\mathcal{F}$ that has been successful for GZSL in video. 
Given a class prototype and data noise, $\mathcal{F}$ is trained to create features of that class that resemble those of the model $\mathcal{M}$. 

\subsection{Training}
\label{sec:training}

\begin{figure*}[t]
    \centering
    \includegraphics[width=\linewidth]{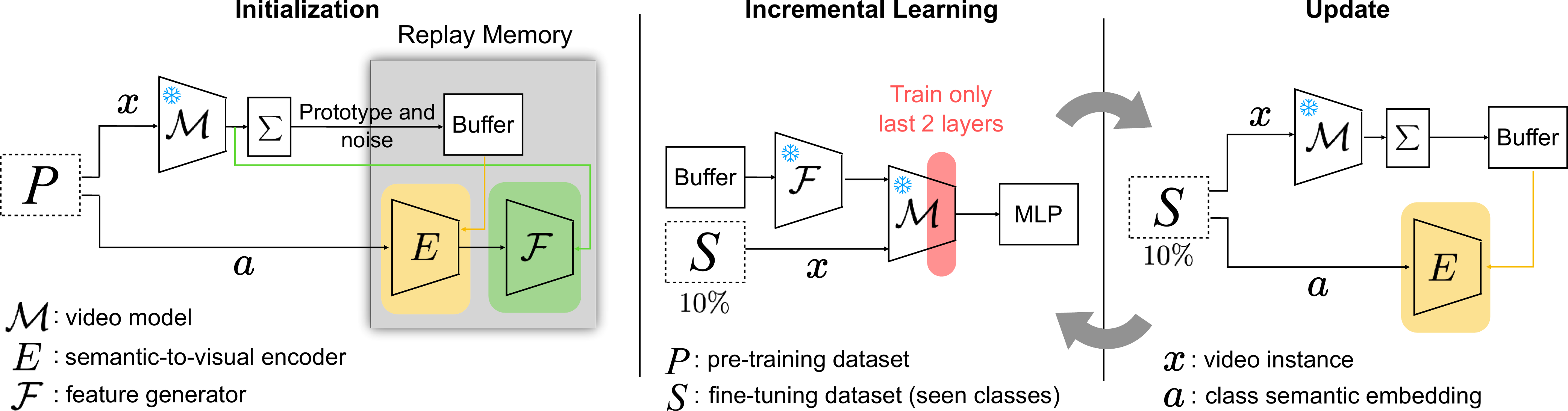}
    \caption{Overview of GIL. The initialization stage involves booting the replay memory, i.e., storing pre-training class prototypes and noise obtained averaging video features encoded by $\mathcal{M}$. $E$ is then trained to produce prototypes given a semantic embedding, and $\mathcal{F}$ is trained to generate visual features from the output of $E$. In the incremental learning stage we fine-tune $\mathcal{M}$ with a mix of synthetic 
    and real features. Synthetic features are generated from the memory buffer, while real features are obtained with the backbone of $\mathcal{M}$. Real features are added gradually sampling a subset of new classes at the time. In the update stage we add prototypes and noise for the new classes to the buffer and fine-tune $E$ with this new data.}
    \label{fig:overview}
\end{figure*}

Our training pipeline can be seen in Figure~\ref{fig:overview}. 
GIL consists of three phases: an Initialization Phase that establishes the foundation model, an Incremental Phase that iteratively introduces new class data, and an Update Phase ensuring the model remains rooted in its previous knowledge while learning new classes. 
\\

\noindent{\bf Initialization Stage.} The foundation model $\mathcal{M}$ starts off as X-Florence~\cite{florence} pre-trained on Kinetics~\footnote{The model is pre-trained on Kinetics400 for the evaluation on Kinetics220, and on Kinetics605 for the evaluation on UCF and HMDB.}.  
We freeze $\mathcal{M}$ and use it to generate visual features for all instances belonging to each class $i$ in the pre-training dataset. For each class $i$ we then calculate its prototype and noise, as described in Sec.~\ref{subsec:GIL}. This is denoted in Fig.~\ref{fig:overview} as the module $\sum$. 
We store prototypes and noise $\mu_i, \sigma_i$ of all classes in the Replay Memory buffer. 
We then train the CVAE ($E$) to predict a prototype and noise $(\hat{\mu_i}, \hat{\sigma_i})$ given a semantic embedding (Stories\cite{stories} in our case). After this, we train the feature generation network $\mathcal{F}$ to generate features for a class given its prototype and noise. These features should be similar to those encoded by $\mathcal{M}$. $\mathcal{F}$ and $E$ are trained from scratch. We use only $\mathcal{F}$ after the initial training of the feature generator network and hence drop $\mathcal{G}$ and $\mathcal{H}$ from the notation and figures for clarity. $\mathcal{F}$ is optimized as described in Eq~\ref{eq:overall}. $E$ is optimized using the mean squared error.
After this stage, $\mathcal{F}$ is permanently fixed. 
\\

\noindent{\bf Incremental Learning Stage.} This stage is the core of our approach and is our contribution to incorporate CL to boost generalization in ZSL. In this stage we fine-tune the last two layers of $\mathcal{M}$ both with synthetic features generated from $\mathcal{F}$ as well as with real features from novel classes. The real features are generated using the backbone of $\mathcal{M}$, sampling videos belonging to a small subset of new classes at every iteration, which we set to be 10\% of the total number of classes in the fine-tuning dataset. The synthetic features are generated using all the prototypes in the memory buffer. In the first iteration, the synthetic features will be only from the pre-training classes. In a subsequent iteration, they will be both from the pre-training classes and the new classes sampled until then. 
The gradual incorporation of new classes helps maintain the generalization of $\mathcal{M}$. 
More formally, at iteration $t$ we generate features $\hat{f_j^k}$ for each class $k$ in the buffer seen during previous stages, which we denote with $C_{t-1}$: 
\begin{equation}
    \hat{f_j^k} = \mathcal{F}(\mu_k, \sigma_k), \forall k\in C_{t-1}, \forall j \in 1\ldots J 
\end{equation}
where $J$ is the number of generated features for class $k$, which we choose to be in the range of the average number of samples in the new set of classes. 
Real features ${f_j^k}$ are computed for the new classes in $C_t$ using the backbone of $\mathcal{M}$: 
\begin{equation}
    f_j^k = \mathcal{M}(x_j), \forall k\in C_{t}, \forall j \in |C_{t}(k)| 
\end{equation}
where $x_j$ is a video from a new class $k$ and $|C_{t}(k)|$ is the number of samples in the class. $\mathcal{M}$ is fine-tuned on both $\hat{f_j^k}$ and $f_j^k$ to predict their class labels. We attach a multi-layer perceptron classifier to fine-tune $\mathcal{M}$, using the standard cross-entropy loss. This classifier is used only to fine-tune $\mathcal{M}$, i.e., we use the output of $\mathcal{M}$ with a nearest neighbor search for prediction during inference, not the output of the classifier (more details on testing later). \\

\noindent{\bf Update Stage.} Once the model has been trained on the new set of classes $C_t$, an update process ensures a smooth transition to the next iteration. The new class prototypes and their noise structure 
are first archived into the Memory Buffer. 
$E$ is then fine-tuned to generate prototypes and noise for the new classes, while $\mathcal{F}$ remains unchanged. 
The iterative process of incremental learning and memory update continues until all training (seen) classes are incorporated in the buffer. 

\subsection{Testing}

\begin{figure}[t]
    \centering
\includegraphics[width=\textwidth]{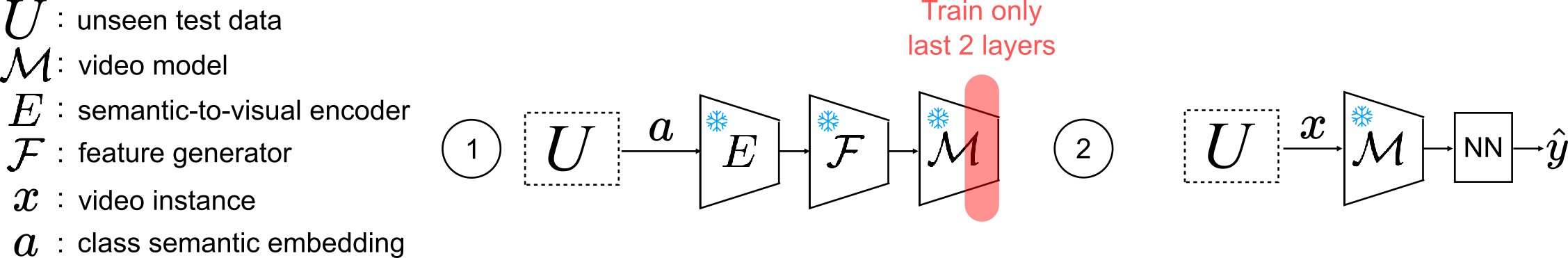}
    \caption{Testing pipeline. (1) We project unseen classes fine-tuning $\mathcal{M}$ with synthetic features. This is done by first feeding $E$ a class semantic embedding. $E$ then outputs a class prototype and noise, which are fed to $\mathcal{F}$ to generate the visual features we use to fine-tune $\mathcal{M}$. (2) After fine-tuning, given a test video instance we perform a nearest neighbor (NN) search to predict class $\hat{y}$.}
    \label{fig:test_gil}
\end{figure}

After training the model on all seen classes, we feed the semantic embeddings of the unseen classes $U$ to $E$. This gives us class prototypes and noise, which we feed to $\mathcal{F}$. This in turn generates unseen class features that are used to fine-tune $\mathcal{M}$ with the additional classifier we employ during the incremental learning stage. We fine-tune $\mathcal{M}$ only on the unseen classes in $U$ in the ZSL setting and on both seen and unseen classes {$U \cup S$} in GZSL. This is depicted in Figure~\ref{fig:test_gil}. We highlight that we do not use any test video instance for fine-tuning. We only use class semantic embeddings (to generate visual features), which is common practice in ZSL~\cite{claster,od,e2e}. At test time, a video sample $x_i$ from an unseen class is fed to $\mathcal{M}$ to produce an embedding. We assign $x_i$ the class of the nearest neighbor ($K$-NN with $K=1$) in the embedding space.

\section{Experimental Analysis}

In this section we first describe the datasets we use in our experiments. We then measure catastrophic forgetting in the standard zero-shot setting, which shows the loss of generalization that happens during the fine-tuning process. We also compare GIL to the current state-of-the-art in the ZSL and GZSL settings. We observe that GIL improves by a wide margin, with up to 20\% improvement with respect to previously published work. We finally present several ablation studies. 
Implementation details can be found in the supplementary material.\\

\noindent{\bf Implementation Details.} The framework for X-CLIP and X-Florence~\cite{florence} consists of three main components: a cross-frame communication transformer, a multi-frame integration transformer, and a text encoder. We offer three CLIP variants: 

\begin{itemize}
    \item X-CLIP-B/32, with 12 layers (\(L_c\)), 12 attention heads (\(N_h\)), embedding dimension of size 768 (\(d\)), and patch size 32(\(p\));
    \item X-CLIP-B/16, with 12 layers (\(L_c\)), 12 attention heads (\(N_h\)), embedding dimension of size 768, and patch size 16(\(p\));
    \item X-CLIP-L/14, with 24 layers (\(L_c\)), 12 attention heads (\(N_h\)), embedding dimension of size 1024 (\(d\)), and patch size 14 (\(p\)).
\end{itemize}

%having 12, 12 and 24 layers (\(L_c\)), 12, 12 and 12 attention heads (\(N_h\)), 768, 768 and 1024 embedding dimension (\(d\)), and patch size 32, 16 and 14 (\(p\)) respectively. 
All models use a 1-layer multi-frame integration transformer. %for all three. 
For X-CLIP the text encoder is the same as in CLIP~\cite{clip}, while for X-Florence it is the same as in Florence~\cite{imgflorence}. 
For X-Florence, the cross-frame communication transformer is replaced with CoSwin-H visual encoder, and a 4-layer multi-frame integration transformer is added. Both X-CLIP and X-Florence utilize a video-specific prompting mechanism set to 2 blocks. We add 2 fully connected layers at the end of the video encoder to convert this to a classification model. The first layer is of size 4096 and the second is equivalent to the dimensions of the semantic vector (768). All hyperparameters are set as in X-Florence~\cite{florence} to ensure fair comparison. 

%The replay memory consists of a memory buffer that has a capacity to store vectors for the pre-training and fine-tuning (seen) classes. 
The CVAE has an encoder that consists of three fully connected layers. It takes in input a vector of size equivalent to the output of the text encoder. %CLIP or Florence depending on the model we use. 
We use a latent dimension of size 512 and use a three-layer decoder for reconstruction.

Our feature generation network follows earlier work~\cite{od,clswgan}. The generator consists of three fully connected layers, where the output layer has dimension matching the video features dimension. The decoder has also three fully connected layers, but its output size corresponds to the class-embedding size. The discriminator has two fully connected layers and outputs a single value. All hidden layers in all networks have size 4096.

\noindent{\bf Datasets and Evaluation Metric.} We conduct experiments on the HMDB-51 \cite{hmdb} and UCF-101 \cite{ucf101} datasets, which are standard choices in zero-shot action recognition. 
These datasets comprise 6,766 and 13,320 videos and encompass 51 and 101 classes, respectively.
We follow the widely adopted 50/50 splits introduced by Xu et al. \cite{xu2017transductive}, where 50\% of the classes are designated as seen, and the other 50\% are designated as unseen. All models are pre-trained on the non-overlapping classes~\cite{e2e} of Kinetics-400~\cite{i3d}.
Since we pre-train our model to avoid overlap to these datasets, the overlap issues that zero-shot action recognition may suffer from~\cite{truze} is not a concern.
We also investigate the Kinetics-600 split \cite{i3d} proposed in ER \cite{er}. In this configuration, the 200 classes from Kinetics-600 \cite{kinetics600} are treated as unseen classes for a model trained on the Kinetics-400 dataset.
Unless otherwise stated, following previous work \cite{zhu2018towards,qin2017zero,stories} we report the average top-1 accuracy and standard deviation from 10 independent runs.

\subsection{Measuring Catastrophic Forgetting in Zero-Shot Learning}

We first show the problem of catastrophic forgetting from a quantifiable point of view. We consider four backbones: X-CLIP-B/16~\cite{florence}, X-CLIP-B/32~\cite{florence}, X-Florence\cite{florence} and Vita-CLIP-B/16~\cite{vita-clip}. We pre-train the models on Kinetics and fine-tune them on UCF-101 with and without GIL. When not using GIL we follow
a standard zero-shot approach based on a feature generation network~\cite{clswgan}, i.e., we remove the replay memory and the incremental learning cycle during training, while the video backbone and the feature generator remain the same (more details in the supplementary material). 
We also evaluate the models on the pre-training dataset (Kinetics), as reported in Table~\ref{tab:cat_for}.
While we see a large improvement on the zero-shot setting with GIL (up to 6.2\%), we see an even larger improvement when testing on the pre-training dataset (up to 24.7\%). We make the following observations from these results. First, there is a clear problem of catastrophic forgetting where the model overfits to the seen classes after fine-tuning (performance without GIL on Kinetics is poor). Second, CL mitigates this problem (performance with GIL is good on Kinetics). Third, features generated with GIL generalize better and do not overfit to the seen classes, as GIL also improves ZSL performance on UCF-101.

\begin{table}[t]
\caption{Evaluating catastrophic forgetting in ZSL. Results indicate top-1 accuracy from a single run. Models were pre-trained on Kinetics-400, fine-tuned on UCF-101, and tested on both UCF-101 and Kinetics. We compare a standard zero-shot approach based on a feature generation network~\cite{clswgan} to our GIL approach.}
\centering
\begin{tabular}{lccc}
\toprule
Model & GIL  & Zero-shot (UCF-101) & Pre-training (Kinetics) \\
\midrule
X-Clip B/16~\cite{florence} & $\times$ & 72.0 & 50.2 \\
X-Clip B/16~\cite{florence} & $\checkmark$ & 77.3 & 71.8 \\
\midrule
X-Clip B/32~\cite{florence} & $\times$ & 70.8 & 49.5 \\
X-Clip B/32~\cite{florence} & $\checkmark$ & 76.1 & 71.1 \\
\midrule
X-Florence~\cite{florence} & $\times$ & 73.2 & 53.7 \\
X-Florence~\cite{florence} & $\checkmark$ & 79.4 & 78.5 \\
\midrule
Vita-CLIP~\cite{vita-clip} & $\times$ & 75.0 & 51.8\\
Vita-CLIP~\cite{vita-clip} & $\checkmark$ & 77.9 & 73.3 \\
\bottomrule
\end{tabular}
\label{tab:cat_for}
\end{table}

\subsection{Comparison to State-Of-The-Art}

\noindent{\bf Zero-shot setting.} We compare to recent state-of-the-art (SOTA) models on the ZSL setting where the evaluation is done only on the unseen classes. 
Recent SOTA use vision-language training, however we also list earlier vision-only methods for completeness. Results on UCF-101 and HMDB-51 are reported in Table~\ref{tab:zsl}, while Table~\ref{tab:zslk600} reports results obtained on Kinetics200 setting~\cite{er}. We note that our method attains new state-of-the-art. We also test popular image-based zero-shot CL frameworks~\cite{cl4,gautam2022tf} using the same backbone and semantic embeddings and see GIL improves by 8-10\%.
\begin{table}[t]
\caption{Comparing GIL to SOTA in the zero-shot setting on HMDB-51 and UCF-101.}
\centering
\begin{tabular}{lccc}
\toprule
Vision-only Methods & Backbone & HMDB-51 & UCF-101 \\
\midrule
OD~\cite{od} (\textit{CVPR'19}) & I3D & 30.2 $\pm$ 2.7  & 26.9 $\pm$ 2.8 \\
OD + SPOT~\cite{gowda2023synthetic} (\textit{CVPRW'23}) & I3D & 34.4 $\pm$ 2.2 & 40.9 $\pm$ 2.6 \\
GGM~\cite{ggm} (\textit{WACV'19}) & C3D & 20.7 $\pm$ 3.1 & 20.3 $\pm$ 1.9 \\
E2E~\cite{e2e} (\textit{CVPR'20}) & C3D  & 32.7 \phantom{+ 0.0} & 48.0 \phantom{+ 0.0} \\
ER-ZSAR~\cite{er} (\textit{ICCV'21}) & TSM & 35.3 $\pm$ 4.6 & 51.8 $\pm$ 2.9 \\
CLASTER~\cite{claster}~\cite{er} (\textit{ECCV'22})& I3D & 42.6 $\pm$ 2.6 & 52.7 $\pm$ 2.2 \\
JigSawNet~\cite{jigsaw} (\textit{ECCV'22}) & R(2+1)D & 39.3 $\pm$ 3.9 & 56.8 $\pm$ 2.8 \\
RGSCL~\cite{SHANG2024112283} (\textit{KBS'24}) & I3D & 34.1 $\pm$ 2.9 & 40.2 $\pm$ 3.8 \\
SupVFD~\cite{niu2024superclass} (\textit{ExSys'24}) & I3D & 31.9 $\pm$ 3.2 & 39.6 $\pm$ 2.9 \\
\toprule
Vision-Language Methods & Backbone & HMDB-51 & UCF-101  \\ \midrule
A5~\cite{a5} (\textit{ECCV'22}) & CLIP-B/16  & 44.3 $\pm$ 2.2 & 69.3 $\pm$ 4.2 \\
X-CLIP-B/16~\cite{florence} (\textit{ECCV'22})& CLIP-B/16 & 44.6 $\pm$ 5.2 & 72.0 $\pm$ 2.3 \\
Vita-CLIP-B/16~\cite{vita-clip} (\textit{CVPR'23})& CLIP-B/16 & 48.6 $\pm$ 0.6 & 75.0 $\pm$ 0.6 \\
EPK-CLIP~\cite{yang2024epk} (\textit{ExSys'24}) & CLIP-B/16 & 48.7 $\pm$ 0.7 & 75.3 $\pm$ 0.9 \\
VicTR~\cite{kahatapitiya2024victr} (\textit{CVPR'24}) & CLIP-B/16 & 51.0 $\pm$ 1.3 & 72.4 $\pm$ 0.3 \\
M$^2$-CLIP~\cite{wang2024multimodal} (\textit{AAAI'24}) & CLIP-B/16 & 47.1 $\pm$ 0.4 & 78.7 $\pm$ 1.2 \\
\textbf{GIL (Ours)} & CLIP-B/16 & 52.7 $\pm$ 1.1 & 77.9 $\pm$ 1.4 \\ \midrule
Tf-gczsl~\cite{gautam2022tf} & Florence & 45.4 $\pm$ 3.5 & 68.6 $\pm$ 2.6 \\
GRCZSL~\cite{cl4} & Florence & 45.9 $\pm$ 3.1 & 69.8 $\pm$ 3.9 \\
X-Florence~\cite{florence} (\textit{ECCV'22}) & Florence & 48.4 $\pm$ 4.9 & 73.2 $\pm$ 4.2 \\
SDR~\cite{stories} (\textit{Arxiv'23}) & Florence & 52.7 $\pm$ 3.4 & 75.5 $\pm$ 3.2 \\
\textbf{GIL (Ours)} & Florence & \textbf{53.9 $\pm$ 1.4} & \textbf{79.4 $\pm$ 1.4} \\
\bottomrule
\end{tabular}
\label{tab:zsl}
\end{table}
In Table~\ref{tab:gzsl} we also compare to recent SOTA on the GZSL setting, where the evaluation is done on both seen and unseen classes. This is a less explored but much more challenging scenario. We see improvements of up to 19.7\% in this setting. These results show that GIL attains new state-of-the-art. In ZSL we see that GIL learns features that generalize better and boost zero-shot performance, while in GZSL we note that the model also better retains pre-training knowledge thanks to the CL paradigm.

\begin{table}[t]
\caption{Comparing GIL to SOTA in the generalized zero-shot setting on HMDB-51 and UCF-101. Reported values show the harmonic mean of the seen and unseen accuracies over 10 runs, $\pm$ $std$. More details in the supplementary.}
\centering
\begin{tabular}{lccc}
\toprule
Vision-only Methods & Backbone & HMDB-51 & UCF-101 \\
\midrule
WGAN~\cite{clswgan} (\textit{CVPR'18}) & I3D & 32.7 $\pm$ 3.4  & 44.4 $\pm$ 3.0 \\
OD~\cite{od} (\textit{CVPR'19}) & I3D & 36.1 $\pm$ 2.2  & 49.4 $\pm$ 2.4 \\
GGM~\cite{ggm} (\textit{WACV'19}) & C3D & 20.1 $\pm$ 2.1  & 23.7 $\pm$ 1.2\\
CLASTER~\cite{claster}~\cite{er} (\textit{ECCV'22})& I3D & 50.8 $\pm$ 2.8  & 52.8 $\pm$ 2.1 \\
RGSCL~\cite{SHANG2024112283} (\textit{KBS'24}) & I3D & 37.6 $\pm$ 2.4 & 50.0 $\pm$ 3.0 \\
SupVFD~\cite{niu2024superclass} (\textit{ExSys'24}) & I3D & 37.0 $\pm$ 2.2 & 50.2 $\pm$ 2.8 \\
FR-VAEGAN~\cite{huang2024generalised} (\textit{Access'24}) & I3D & 38.6 $\pm$ 0.1 & 54.7 $\pm$ 0.2\\
\toprule
Vision-Language Methods & Backbone & HMDB-51 & UCF-101 \\ \midrule
SDR~\cite{stories} (\textit{ArXiv'23}) & Florence & 53.5 $\pm$ 3.3 & 57.8 $\pm$ 4.1 \\
\textbf{GIL (Ours)} & Florence & \textbf{55.1 $\pm$ 1.9} & \textbf{77.5 $\pm$ 1.9} \\
\bottomrule
\end{tabular}
\label{tab:gzsl}
\end{table}

\begin{table}[t]
\caption{Comparing GIL to SOTA in the zero-shot setting on Kinetics-200. Following ER \cite{er}, seen classes are those in Kinetics-400, while the new 200 classes in Kinetics-600 are unseen. Results report average top-1/5 accuracy ± $std$ from 10 runs.}
\centering
\begin{tabular}{lccc}
\toprule
Vision-only Methods & Backbone & Top-1  & Top-5\\
\midrule
SJE~\cite{sje} (\textit{CVPR'15}) & TSM & 22.3 $\pm$ 0.6 & 48.2 $\pm$ 0.4 \\
ER-ZSAR~\cite{er} (\textit{ICCV'21}) & TSM & 42.1 $\pm$ 1.4 & 73.1 $\pm$ 0.3 \\
JigSawNet~\cite{jigsaw} (\textit{ECCV'22}) & R(2+1)D & 45.9 $\pm$ 1.6 & 78.8 $\pm$ 1.0 \\
SDR-I3D~\cite{stories} (\textit{ArXiv'23}) & I3D & 50.8 $\pm$ 1.9 & 82.9 $\pm$ 1.3 \\
\toprule
Vision-Language Methods & Backbone & Top-1 & Top-5 \\ \midrule
X-CLIP-B/16~\cite{florence} (\textit{ECCV'22}) & CLIP B/16 & 65.2 $\pm$ 0.4 & 86.1 $\pm$ 0.8 \\
Vita-CLIP-B/16~\cite{vita-clip} (\textit{CVPR'23}) & CLIP B/16 & 67.4 $\pm$ 0.5 & 86.9 $\pm$ 0.6 \\
\textbf{GIL (Ours)} & CLIP B/16 & 70.9 $\pm$ 1.3 & 90.7 $\pm$ 1.2 \\ \midrule
SDR~\cite{stories} (\textit{ArXiv'23}) & Florence & 55.1  $\pm$ 2.2 & 86.1 $\pm$ 3.1  \\
X-Florence~\cite{florence} (\textit{ECCV'22}) & Florence & 68.8 $\pm$ 0.9 & 88.4 $\pm$ 0.6 \\
\textbf{GIL (Ours)} & Florence & \textbf{72.5 $\pm$ 1.6} & \textbf{93.1 $\pm$ 1.1} \\
\bottomrule
\end{tabular}
\label{tab:zslk600}
\end{table}

\subsection{Ablation Study}

\noindent{\bf Effect of Replay Memory.} 
In Table~\ref{tab:abl_rm} we explore different design choices for the replay memory.   
First, we remove the memory altogether. Second, we store some samples for each class chosen randomly instead of storing the class prototype and standard deviation. Third, we replace the standard deviation with a few more random samples (shown in the table as Class Prototype + Random Samples). All options improve over the baseline of not using a memory replay. We also observe that the choice of class representation has a strong effect on the results, and using the proposed class prototype and data noise is the most effective of all. Storing the class distribution as noise also significantly lowers the standard deviation in the reported top-1 accuracy, i.e., results obtained with this choice of replay memory are more robust as they vary less. 

\begin{minipage}{0.45\linewidth}
\captionof{table}{Comparing different choices of replay memory in the zero-shot setting. No Mem: No Replay Memory, Rand: Random Sampling, Proto + Rand: Class Prototype + Random Sampling, Proto + Noise: Class Prototype + Data Noise. }
\resizebox{\columnwidth}{!}{
\noindent
\begin{tabular}{@{}lcc}
\toprule
Method & HMDB-51 & UCF-101 \\
\midrule
No Mem & 48.7 $\pm$ 4.4 & 73.5 $\pm$ 3.5 \\
Rand & 49.4 $\pm$ 4.1 & 75.1 $\pm$ 2.9 \\
Proto + Rand & 50.6 $\pm$ 3.8 & 76.6 $\pm$ 2.7 \\
Proto + Noise & \textbf{53.9 $\pm$ 1.4} & \textbf{79.4 $\pm$ 1.4} \\
\bottomrule
\end{tabular}
}
\label{tab:abl_rm}
\end{minipage}
\hfill
\begin{minipage}{0.47\linewidth}
\captionof{table}{Comparing different backbones in the zero-shot setting. CLIP~\cite{florence}: X-CLIP-B/16 (\textit{ECCV'22}), X-F~\cite{florence}: X-Florence~\cite{florence} (\textit{ECCV'22}), Vita~\cite{vita-clip}: Vita-CLIP-B/16~\cite{vita-clip}(\textit{CVPR'23})}
\centering
\resizebox{\columnwidth}{!}{
\begin{tabular}{lcc}
\toprule
Backbone & HMDB-51 & UCF-101 \\
\midrule
CLIP & 44.6 $\pm$ 5.2 & 72.0 $\pm$ 2.3 \\
CLIP + \textbf{GIL} & 48.9 $\pm$ 1.9 & 77.3 $\pm$ 1.5 \\
\midrule
X-F & 48.4 $\pm$ 4.9 & 73.2 $\pm$ 4.2 \\
X-F + \textbf{GIL} & \textbf{53.9 $\pm$ 1.4} & \textbf{79.4 $\pm$ 1.4} \\
\midrule
Vita & 48.6 $\pm$ 0.6 & 75.0 $\pm$ 0.6 \\
Vita + \textbf{GIL} & 52.7 $\pm$ 1.1 & 77.9 $\pm$ 1.4 \\
\bottomrule
\end{tabular}
}
\label{tab:abl_bb}
\end{minipage} \\

\noindent{\bf Choice of Backbone.}
\label{sec:abl_model}
GIL is model-agnostic. We chose X-Florence~\cite{florence} due to its strong performance on zero shot learning, 
and explore now the effect of using other backbones. Results can be seen in Table~\ref{tab:abl_bb}. We note that all backbones improve when trained with GIL, which highlights its generalizability. Since we do not make any specific assumptions on the backbone architecture, we believe GIL is a general framework that can be plugged to any model. \\

\begin{figure}[t]
    \centering
    \includegraphics[width=\textwidth]{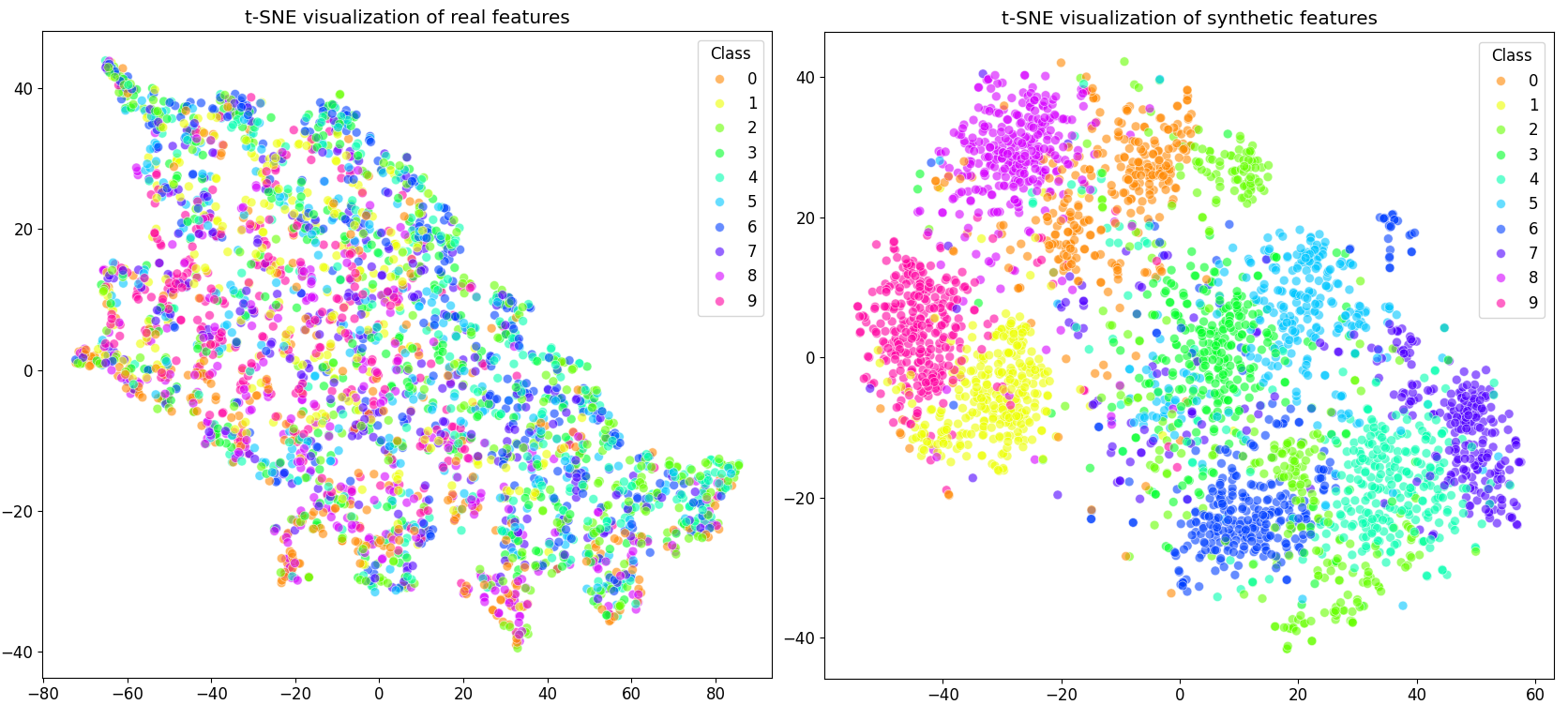}
    \caption{Comparing real (left) vs real (right) features using t-SNE projections. We show the embeddings obtained from 10 random classes in HMDB-51. Synthetic features appear more compact than real features, which is beneficial to train the video model. We believe this is the reason why using exclusively synthetic features for the pre-training classes works better than using real features or a mix of the two.}
    \label{fig:feats}
\end{figure}

\noindent{\bf Using Real Data from Kinetics.} During the incremental learning stage we use both synthetic and real features from the \textit{fine-tuning} dataset, but only use synthetic features from the \textit{pre-training} dataset (generated from the memory). A natural question to ask is thus why not use real data from the pre-training dataset as well during the incremental learning stage. We explore this question and show results with the zero-shot setting in Table~\ref{tab:abl_real}. Here we vary how data from the \textit{pre-training} dataset is obtained, however for the \textit{fine-tuning} dataset we keep using a mix of synthetic and real data. We consider three scenarios. First, we sample only real data during the incremental stage, i.e., we do not use the feature-generating network and instead use real features obtained with the video backbone. 
In the second scenario we use a combination of real and synthesized data (with a 50:50 ratio). The last scenario is GIL's default setting, using purely synthetic data. The baseline in this experiment is the original X-Florence model. Regardless of the type of data we use for the pre-training dataset, the replay memory improves over the baseline significantly, and using purely synthetic data gives the best performance. 
We hypothesize this happens because synthetic features tend to form a more compact representation, which is beneficial to fine-tune the video model. In Figure~\ref{fig:feats} we show this, plotting t-SNE projections~\cite{hinton2008visualizing} of real and synthetic features for a random set of 10 classes in HMDB-51\footnote{We show 10 classes for clarity, but plots were identical with several random samples.}. Indeed, synthetic features form more compact clusters, whereas real features are more scattered. This is not surprising as synthetic features are generated from class prototypes while real features are obtained from video instances.\\

\noindent{\bf Choice of Semantic Embedding.} We use Stories~\cite{stories} as our semantic embedding. Here we compare with other encodings: word2vec~\cite{w2v}, sen2vec~\cite{s2v} and ER~\cite{er}. We report results from this experiment with the zero-shot setting in Table~\ref{tab:abl_sem}. We note that GIL improves over the baseline (the original X-Florence model) regardless of the semantic embedding we use. This further highlights the generalizability of our approach. Results with Word2Vec improve only marginally. We believe this is due to the limited semantic space word2vec is able to encode in comparison to the others. \\

\noindent{\bf Fine-tuning the Feature Generator.} We keep the feature generator fixed after initialization. We do this for two reasons. Firstly, training a feature generator is expensive, and we wish to keep computational cost low. More importantly, we did not find a discernible difference in performance when fine-tuning the feature generator. More results on this in the Supplemental Material. 

\begin{minipage}{0.47\linewidth}
\captionof{table}{Comparing types of data from the pre-training dataset for the incremental stage.}
\centering
\resizebox{\columnwidth}{!}{%
\begin{tabular}{lcc}
\toprule
Type of Data & HMDB-51 & UCF-101 \\
\midrule
Baseline~\cite{florence} & 48.4 $\pm$ 4.9 & 73.2 $\pm$ 4.2 \\
Real Only & 51.8 $\pm$ 2.4 & 76.5 $\pm$ 2.2 \\
Real + Synthetic & 52.7 $\pm$ 1.9 & 78.5 $\pm$ 2.4 \\
Synthetic Only & \textbf{53.9 $\pm$ 1.4} & \textbf{79.4 $\pm$ 1.4} \\
\bottomrule
\end{tabular}
}
\label{tab:abl_real}
\end{minipage}
\hfill
\begin{minipage}{0.43\linewidth}
\captionof{table}{Comparing semantic embeddings for the incremental stage.}
\centering
\resizebox{\columnwidth}{!}{%
\begin{tabular}{lcc}
\toprule
SE & HMDB-51 & UCF-101 \\
\midrule
Baseline~\cite{florence} & 48.4 $\pm$ 4.9 & 73.2 $\pm$ 4.2 \\
Word2Vec~\cite{w2v} & 48.9 $\pm$ 4.1 & 73.9 $\pm$ 4.1 \\
Sen2Vec~\cite{s2v} & 50.8 $\pm$ 3.1 & 76.7 $\pm$ 2.9 \\
ER~\cite{er} & 51.9 $\pm$ 1.5 & 77.9 $\pm$ 1.3 \\
Stories~\cite{stories} & \textbf{53.9 $\pm$ 1.4} & \textbf{79.4 $\pm$ 1.4} \\
\bottomrule
\end{tabular}
}
\label{tab:abl_sem}
\end{minipage}

\section{Detailed Generalized Zero-Shot Action Recognition Results}

In order to better analyze performance of the model on GZSL, we report the average seen and unseen accuracies along with their harmonic mean. The results on the UCF101~\cite{ucf101} and HMDB51~\cite{hmdb} datasets are reported in Table~\ref{tab:gzsl_vid}. Results are averaged from 10 runs, where for each run we create a different random train/test split (which is the same for all models). %The reported results are on the same set of 10 random splits for fair comparison.
We note that in this more challenging setting GIL performs significantly better than previous state-of-the-art (especially on UCF-101), highlighting that the model trained with GIL is able to better retain knowledge and generalize to unseen classes. We do not report on Truze~\cite{truze} as we have no overlap between the train and test classes.

\begin{table}[t]
\begin{center} 
\caption{Generalized Zero-Shot results, where `u', `s' and `H' correspond to average unseen accuracy, average seen accuracy and the harmonic mean of the two. All the reported results are on the same splits.} \label{tab:gzsl_vid}
\begin{tabular}{ *{7}{c} }
\toprule
Model   & \multicolumn{3}{c}{HMDB51}& \multicolumn{3}{c}{UCF-101}\\
\midrule
&  u & s & H & u & s & H \\
\midrule
WGAN \cite{clswgan}  & 23.1 & 55.1 & 32.5 & 20.6 & 73.9 & 32.2\\
OD \cite{od}  & 25.9 & 55.8 & 35.4 & 25.3 & 74.1 & 37.7\\
OD + SPOT \cite{gowda2023synthetic} & 26.7 & 54.1 & 35.7 & 28.3 & 74.1 & 40.9 \\
CLASTER \cite{claster} & 43.7 & 53.3 & 48.0 & 40.8 & 69.3 & 51.3\\
SDR \cite{stories}  & 50.1 & 57.5 & 53.5 & 47.3 & 81.2 & 59.7\\
\textbf{GIL} (Ours) & \textbf{52.8} & \textbf{57.8} & \textbf{55.1} & \textbf{68.2} & \textbf{89.8} & \textbf{77.5} \\
\bottomrule
\end{tabular}
\end{center}
\end{table}

\section{How Much Does Sampling Percentage of Data Affect Model Performance?}

We choose to sample 10\% of data from the seen classes at each iteration. Here we consider a few other settings sampling different amounts of data, namely: 1\%, 5\%, 10\%, 20\%, 50\% and 100\% of data at once. Table~\ref{tab:sampl_abl} shows the results. We do not see a notable change in performance when increasing from 1\% to 10\%, however beyond 10\% the performance starts dropping. The higher the percentage of data we sample, the faster training is. These results suggest the generalization ability of the model are affected when it is trained ``too fast'', and that a gradual and slow introduction of new data is accordingly beneficial. % , and . would be our training and hence we fix this at 10\%.

\begin{table}[t]
\centering
\caption{Evaluating the impact of sampling percentage of data on zero-shot performance.}
\label{tab:sampl_abl}
\begin{tabular}{lcc}
\toprule
\% of Data  & HMDB51 & UCF101 \\
\midrule
1  & 53.7 $\pm$ 1.3 & 79.1 $\pm$ 1.6 \\
5 & \textbf{53.9} $\pm$ \textbf{1.1} & 79.2 $\pm$ 1.5 \\
10 & \textbf{53.9} $\pm$ \textbf{1.4} & \textbf{79.4} $\pm$ \textbf{1.4} \\
20 & 53.2 $\pm$ 1.5 & 78.6 $\pm$ 1.0 \\
50 & 49.7 $\pm$ 2.8 & 75.2 $\pm$ 2.6 \\
100 & 48.6 $\pm$ 3.7 & 73.1 $\pm$ 3.4 \\
\bottomrule
\end{tabular}

\end{table}

\section{Using the Text Encoder of X-Florence Directly}

Instead of using any semantic embedding, we could potentially leverage the text encoder from X-Florence directly. We try this and report these results in Table~\ref{tab:abl_te}. We see that even using a simple Word2Vec embedding does better than using the text encoder to produce semantic embeddings.

\begin{table}[t]
\centering

\caption{Using the text encoder from the base model as semantic embedding, compared to the other semantic embeddings we evaluate in this work.}
\label{tab:abl_te}
\begin{tabular}{lcc}
\toprule
Semantic Embedding & HMDB-51 & UCF-101 \\
\midrule
X-Florence Text Encoder & 47.2 $\pm$ 4.1 & 71.6 $\pm$ 3.8 \\
Word2Vec & 48.9 $\pm$ 4.1 & 73.9 $\pm$ 4.1 \\
Sen2Vec & 50.8 $\pm$ 3.1 & 76.7 $\pm$ 2.9 \\
ER & 51.9 $\pm$ 1.5 & 77.9 $\pm$ 1.3 \\
Stories & \textbf{53.9 $\pm$ 1.4} & \textbf{79.4 $\pm$ 1.4} \\
\bottomrule
\end{tabular}
\end{table}

\section{How Much Does the Number of Generated Synthetic Samples Affect Model Performance?}

We generate synthetic samples from already seen classes in order to train our classifier with synthetic features and refresh its memory. All experiments in the main paper were conducted ensuring that the distribution of the newly sampled classes and the synthetic features were similar, i.e., we generate a number of synthetic features roughly equal to the number of new real samples. We also experimented with generating fewer samples (20\% to 70\% of the average number of samples in the new classes) and more samples (150\% to 200\%). We report these results in Table~\ref{tab:abl_num_syn}, where 100\% corresponds to what we do for all results in the main paper. We see that either over generating or under generating leads to poorer results, which suggests that keeping the number of synthetic and real features balanced is beneficial to the model.

\begin{table}[t]
\centering
\caption{Using different percentages of synthetic data (relative to the size of the new classes) to train the classifier.}
\label{tab:abl_num_syn}
\begin{tabular}{lcc}
\toprule
Percentage of synthetic samples & HMDB-51 & UCF-101 \\
\midrule
20 & 48.4 $\pm$ 2.5 & 74.2 $\pm$ 1.6 \\
50 & 50.5 $\pm$ 1.9 & 76.1 $\pm$ 1.8 \\
70 & 52.9 $\pm$ 1.8 & 78.6 $\pm$ 1.5 \\
100 & \textbf{53.9 $\pm$ 1.4} & \textbf{79.4 $\pm$ 1.4} \\
120 & 53.6 $\pm$ 1.7 & 77.9 $\pm$ 1.8 \\
150 & 52.7 $\pm$ 1.6 & 77.2 $\pm$ 1.6 \\
200 & 49.7 $\pm$ 1.3 & 76.4 $\pm$ 1.5 \\
\bottomrule
\end{tabular}
\end{table}

\section{Freezing the Feature Generator}

As mentioned in the paper we freeze the feature generator after it is trained on the pre-training dataset. We do this both to save computation resources and because in practice freezing this models leads to better results. We show this comparing results obtained fine-tuning the feature generator on the seen classes (one single fine-tuning) and fine-tuning the feature generator in the incremental learning stage together with the other models. Results are reported in Table~\ref{tab:freeze}, where we see that indeed freezing the feature generator gives better results. We speculate this is the case because fine-tuning the generator makes the overall framework more difficult to optimize, i.e., it is easier to optimize the video model with features generated from a stable generator.

\begin{table}[t]
\centering
\caption{Comparing results obtained fine-tuning and freezing the feature generator.}
\label{tab:freeze}
\begin{tabular}{lcc}
\toprule
Fine-tune stage & HMDB-51 & UCF-101 \\
\midrule
Fine-tune on seen classes & 49.5 $\pm$ 1.9 & 75.2 $\pm$ 1.6 \\
Incremental fine-tune & 53.1 $\pm$ 1.2 & 77.9 $\pm$ 1.3 \\
Frozen & \textbf{53.9 $\pm$ 1.4} & \textbf{79.4 $\pm$ 1.4} \\
\bottomrule
\end{tabular}
\end{table}

\section{Experiments with CL in ZSL setup}
\begin{figure}
    \centering
    \includegraphics[width=0.85\columnwidth]{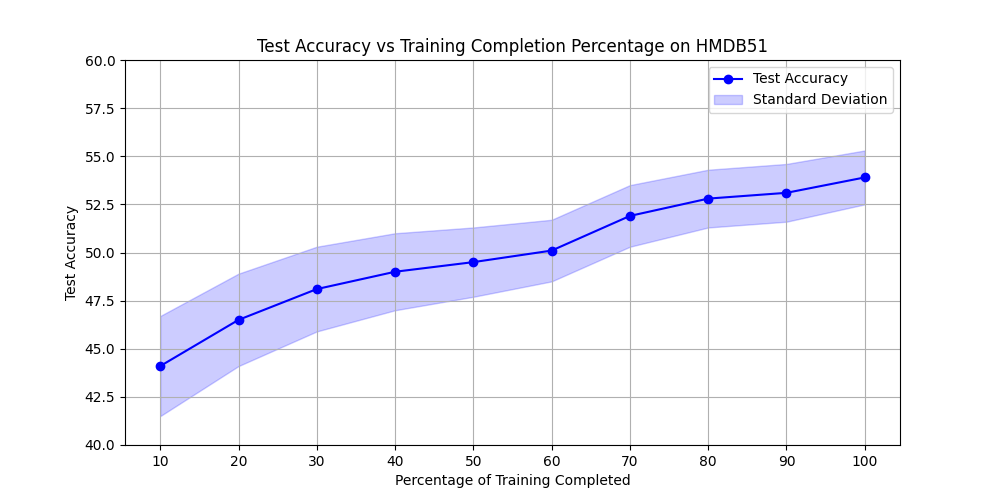}
    \vspace{-10pt}
    \caption{Test accuracy versus training completion percentage.\vspace{-20pt}} %, showing the model's performance with 10\% incremental training stages.} %The shaded area represents the standard deviation, indicating the variability in test accuracy across different stages of training.}
    \label{fig:cont}
\end{figure}

All our experiments are in the ZSL (or generalized ZSL) setting, where test classes are disjoint with the training set at all times. We also evaluate here our model after each fine-tuning stage, i.e.,  after each time we introduce the 10\% of new classes. As expected, performance grows steadily as we introduce more data as seen in Fig~\ref{fig:cont}. This also justifies the idea of slowly introducing data as opposed to directly fine-tune using all data.

\section{Conclusion}

This paper introduces Generative Iterative Learning (GIL), a method that combines a generative approach for zero-shot learning within a continual learning paradigm. This work is the first to use CL to improve zero-shot action recognition in video. 
The core innovations of GIL lie in its replay memory and incremental learning. The memory stores class prototypes that help synthesize features from both past classes and new, real examples. The continual updating of this memory enables the retraining of the classification model in a balanced manner, ensuring equal exposure to old and new information. Experiments show that GIL 
significantly boosts the model's generalization capabilities in unseen classes. Results demonstrate we achieve a new state-of-the-art in zero-shot action recognition on three standard benchmarks, irrespective of the video backbone we use. 

\noindent{\bf Limitations.} One of the possible limitations of GIL is that relying on synthesized features could introduce biases or inaccuracies in feature representation, potentially affecting the model's performance on highly varied or dynamic datasets. Further, evaluating on datasets with viewpoints that differ from those seen in training may not benefit from the proposed approach.

\noindent{\bf Future Work.} While this work has shown great promise in the integration of a continual learning framework for action recognition in the zero-shot setting, the idea could potentially be tested on few-shot and semi-supervised settings as the problem of catastrophic forgetting would still exist. Further, GIL could also be applied for other video tasks such as video localization. 

% ---- Bibliography ----
%
% BibTeX users should specify bibliography style 'splncs04'.
% References will then be sorted and formatted in the correct style.
%
\bibliographystyle{splncs04}
\bibliography{main}
\end{document}